\documentclass[conference]{IEEEconf}

\input epsf
\usepackage{graphicx}
\usepackage{multirow}
\usepackage{titlesec}
\usepackage{balance} 
\usepackage{stfloats}
\usepackage{xcolor}
\usepackage{tcolorbox, xcolor}
\usepackage[table,dvipsnames]{xcolor}
\usepackage[ruled,vlined,linesnumbered]{algorithm2e}
\usepackage{booktabs}
\usepackage{threeparttable}
\usepackage{minted}
\usepackage{tabularx}
\usepackage{multirow}
\usepackage{array}
\usepackage{url}
\usepackage{listings}
\usepackage{float}
\usepackage{amsmath}
\renewcommand\thesection{\arabic{section}} % arabic numerals for the sections

\renewcommand\thesubsectiondis{\thesection.\arabic{subsection}}% arabic numerals for the subsections

\renewcommand\thesubsubsectiondis{\thesubsectiondis.\arabic{subsubsection}}% arabic numerals for the subsubsections

\AtBeginDocument{%
  }
\begin{document}

\title{\textbf{\Large AutoACSL: Synthesizing ACSL Specifications by Integrating LLMs with CPG-Based Static Analysis \\}}

\author{Han Zhou$^{1}$, Yu Luo$^{2}$, and Dianxiang Xu$^{1}$\\
	\normalsize $^{1}$University of Missouri--Kansas City, Kansas City, USA\\
	\normalsize $^{2}$University of Central Missouri, Warrensburg, USA\\
	
	\normalsize hzb4f@umsystem.edu, yuluo@ucmo.edu, dxu@umkc.edu\\
}
% \author{Anonymous Authors \\ 
%  Anonymous Institutions}

% \author{
%   Han Zhou\inst{1} \and
%   Yu Luo\inst{2} \and
%   Dianxiang Xu\inst{1}
% }
% %
% \authorrunning{Zhou et al.}
% \institute{
%   University of Missouri--Kansas City, Kansas City, USA\\
%   \email{\{hzb4f, dxu\}@umsystem.edu}
%   \and
%   University of Central Missouri, Warrensburg, USA\\
%   \email{yuluo@ucmo.edu}
% }
%+++++++++++++++++++++++++++++++++++++++++++

% use only for invited papers
%\specialpapernotice{(Invited Paper)}

% make the title area
\maketitle
\begin{abstract}
Generating formal specifications for C programs remains a challenge in formal verification due to the manual effort, expertise, and semantic precision required. While recent advancements in large language models (LLMs) offer promise in automating specification synthesis, current approaches often lack semantic depth and produce unverifiable or incomplete contracts. To address these limitations, we introduce AutoACSL, a novel framework that integrates LLM prompting with semantic features extracted from Code Property Graphs (CPGs). AutoACSL performs static analyses to extract key semantic elements, including arithmetic operations, loop and recursion structures, and return value propagation, which are encoded into structured prompts. These prompts enable the LLM not only to generate normal behavioral specifications but also to include constraints that prevent inputs leading to runtime errors. AutoACSL employs a feedback-driven synthesis loop, where candidate specifications are verified using Frama-C/WP and refined iteratively until verification succeeds or a termination condition is met. Evaluated on 604 programs drawn from diverse datasets, AutoACSL achieves a 98\% specification generation success ratio and a 96\% full proof ratio when paired with Gemini-3. Compared to a code-only baseline, AutoACSL improves the full proof ratio by 24.7\% to 51.7\% across four LLMs (GPT-o4 Mini, GPT-5.2, Grok-4.1, and Gemini-3), demonstrating that integrating large language models with CPG-based static analysis substantially enhances both generation robustness and verification effectiveness for automated ACSL specification synthesis.
\end{abstract}
\IEEEoverridecommandlockouts
\vspace{1.5ex}
% no keywords

% For peer review papers, you can put extra information on the cover
% page as needed:
% \begin{center} \bfseries EDICS Category: 3-BBND \end{center}
%
% for peerreview papers, inserts a page break and creates the second title.
% Will be ignored for other modes.
\IEEEpeerreviewmaketitle

\section{Introduction}
Formal specification plays a central role in the rigorous verification of software. By precisely defining a program’s expected behavior using preconditions, postconditions, and invariants, specifications serve as the foundation for correctness proofs. However, the widespread adoption of formal verification remains limited in practice—largely due to the manual effort and expertise required to write correct, complete, and verifiable specifications. For instance, authoring reliable ACSL (ANSI/ISO C Specification Language) \cite{frama-c} contracts for C programs demands both deep semantic understanding and familiarity with formal logic and tool support (e.g., Frama-C).

Large Language Models (LLMs) have recently shown promise in automating aspects of specification synthesis \cite{George2025Seeking,George2025Specify,Christian2023,liu2024enhancing}. Studies have demonstrated that prompting models to generate specifications from source code can yield surprisingly plausible outputs. Nevertheless, these LLM-generated specifications often suffer from semantic gaps and verification failures, as models lack access to structured semantic context. Treating source code as unstructured text causes generated specifications to be incorrect or unverifiable — even when they appear syntactically valid. Moreover, LLMs tend to overfit to surface patterns, struggle with semantic reasoning, and fail to enforce safety constraints such as memory validity, integer overflows, or termination conditions in recursive functions.

To address these limitations, we present AutoACSL, a novel framework that combines the generative power of LLMs with semantics-aware static analysis to produce verifiable ACSL specifications, including
preconditions, postconditions, frame conditions, loop contracts, and termination clauses. AutoACSL prompts the LLM using semantic features extracted from the Code Property Graph (CPG) \cite{yamaguchi2014modeling} of the given C program. CPG is a unified program representation combining syntax, control flow, data dependencies, and control dependencies. The CPG analysis identifies such critical information as arithmetic computations, loop and recursion structures, and output propagation paths. Prompts constructed with these features enable the LLM not only to generate normal behavioral specifications, but also to include constraints that prevent inputs leading to runtime errors, such as integer overflows and out-of-bounds accesses. AutoACSL adopts a feedback-driven loop: If the verification of a generated specification fails due to syntax errors or unproven verification conditions, it may reconstruct the prompt using the verification feedback and re-engage the LLM, if necessary, to generate an improved specification.

Evaluation results on a total of 604 C programs drawn from diverse datasets show that AutoACSL, when coupled with Gemini 3.0, successfully generates verifiable specifications for 98\% of the programs, with 96\% of the generated specifications fully proved against the corresponding C code. Our ablation study further demonstrates that CPG-based static analysis and the guided prompting strategy play a critical role in achieving these gains, improving specification generation robustness by up to 64.4\% and increasing the full proof ratio by up to 51.7\% when applied to Grok-4.1, compared to a code-only prompting baseline.

The remainder of this paper is organized as follows. Section 2 reviews related work. To make the paper self-contained, Section 3 provides background information on ACSL, Frama-C, and CPG. Section 4 presents the AutoACSL framework. Section 5 elaborates on feature extraction from the CPG, and Section 6 discusses prompt construction with the extracted features. Section 7 presents our experiment results. Finally, Section 8 concludes the paper.

\section{Related Work}

\subsection{Traditional Specification Synthesis from C Programs}

Traditional approaches to specification synthesis from C programs fall into two main categories: dynamic and static. Dynamic methods infer likely pre/postconditions or invariants based on observed execution behavior. Daikon~\cite{Ernst2007} instruments programs to trace variable values at key locations (e.g., function boundaries) and infers candidate specifications from recurring patterns across executions. However, the resulting specifications are only guaranteed for the observed inputs, which may lead to overfitting. SymInfer~\cite{Nguyen2022} addresses this limitation by combining concrete traces with symbolic states, improving the generalization of inferred properties. DySy~\cite{Csallner2008} further integrates symbolic execution with concrete runs, using counterexamples to iteratively refine candidate invariants. SPEEDY~\cite{Cok_2014} blends lightweight runtime monitoring with Frama-C's static analyzers to suggest specifications on-the-fly as developers write code, offering a balance between automation and user involvement. CLIPER~\cite{lo2007efficient} mines recurring behavioral patterns from execution traces to infer temporal specifications. %It captures iterative patterns across and within traces without window constraints and expresses them as generalized temporal rules.

Static methods analyze code structure and semantics without executing it, aiming for soundness across all possible program paths. Kindspec~\cite{alpuente2020abstract,alpuente2023automated} uses symbolic execution over programs written in the KernelC subset of the C language to synthesize abstract pre/postconditions expressed as logical contracts. Araujo et al.\cite{dordowsky2015experimental} exemplify a manual approach, translating avionics requirements into ACSL annotations and verifying them with Frama-C. AutoDeduct\cite{Amilon2025} automates the contract generation process through symbolic execution and Horn clause reasoning. TriCera~\cite{amilon2021automated} follows a similar route, translating C code into constrained Horn clauses and using SMT solvers to infer ACSL function contracts.

\subsection{ML/LLM-Assisted Specification Synthesis}
%With the rapid advancements in machine learning (ML), particularly in LLMs, there has been growing interest in leveraging ML techniques to automatically generate formal specifications.

\textbf{LLM-based generation for C and ACSL.}
Several studies have used LLMs to synthesize ACSL specifications for C programs. George et al.\cite{George2025Seeking} prompt GPT-4 to generate ACSL contracts directly from C source code, and demonstrate that the model can often distinguish between specifications that capture intended behavior versus those that mirror implementation quirks. Building on this, George et al. \cite{George2025Specify} integrate outputs from Frama-C’s Eva \cite{frama-c-eva} and PathCrawler \cite{frama-c-pathcrawler} into the LLM prompt, showing that symbolic error information and test-based traces help the model generate more correct and non-trivial contracts. AutoSpec~\cite{Cheng2024} leverages static analysis to parse the program and build a call graph, then enters an iterative neuro-symbolic loop in which LLM-generated specification candidates are continually refined based on feedback from formal verification tools. VeCoGen~\cite{sevenhuijsen2025vecogen} takes an ACSL specification, a natural language description, and unit tests as input, then uses an LLM to generate candidate C programs. %It verifies them using Frama-C’s WP and RTE plugins and iteratively refines candidates based on verification feedback.

Janßen et al.\cite{Christian2023} evaluate ChatGPT on 106 small C programs, showing that it can generate correct and useful loop invariants that enable Frama-C and CPAchecker to verify code that would otherwise fail. They identify key failure patterns—such as overly weak or invalid invariants—and suggest feedback-driven refinement. Guangyuan et al.\cite{Guangyuan2024} propose a neuro-symbolic pipeline where GPT-3.5 generates candidate invariants, which are then verified by CBMC; counterexamples are fed back to improve subsequent proposals. ACInv~\cite{liu2024enhancing} extracts loop context through static analysis and prompts an LLM to generate loop invariants, then iteratively refines those invariants via an LLM-based evaluator that strengthens, weakens, or rejects them until they pass verification.

\textbf{LLM-based generation for other languages.}
Beyond C, several researchers have explored LLM-based specification synthesis in Java, C++, Rust, and Solidity. Teuber and Beckert~\cite{Teuber2025} use GPT to complete partial JML annotations for Java methods. Their framework uses the KeY verifier in a loop to repair invalid contracts, enabling automated annotation of real-world codebases. Greiner et al.~\cite{greiner2024automated} use CodeT5-based models to generate JML pre/postconditions for Java methods, producing syntactically and semantically valid specifications from over 14,000 training examples. SpecGen~\cite{Lezhi2025} employs a two-phase strategy: first, using dialogue-style prompting to help an LLM draft pre/postconditions and invariants; second, applying mutation and heuristic selection to repair and verify those contracts. Wang et al.~\cite{wang2025} propose Preguss with a stronger focus on large-scale programs, potential-runtime-error guidance, and interprocedural refinement. SAFE~\cite{chen2024automated} generates function specifications by prompting LLMs with Rust function signatures and doc comments to produce candidate preconditions and postconditions in the Verus specification language. These specifications are iteratively refined through self-evolution. In the object-oriented domain, ClassInvGen~\cite{Chuyue2025} co-generates C++ class invariants and test inputs, using the LLM to propose candidate invariants that are validated through execution. In the smart contract space, Liu et al.~\cite{Ye2025} build PropertyGPT to prompt GPT-4 with retrieved examples from existing Certora reports, generating new formal properties for Solidity programs. These properties are then checked and refined using custom provers, achieving both recall and generalization. 

\section{Preliminaries}

\subsection{C Program Specifications in ACSL}
The main constructs of ACSL specifications are as follow:

\begin{itemize}
    \item \textbf{Preconditions (\texttt{requires}):}  A precondition specifies a set of conditions that must hold true before a function is invoked. In ACSL, preconditions are specified using the requires keyword within the annotation block.
    
    \item \textbf{Postconditions (\texttt{ensures}):} Postconditions describe properties to be hold upon function return. They may refer to the return value (\texttt{\textbackslash result}) and to the values of expressions in the pre-state (\texttt{\textbackslash old(e)}). 
    
    \item \textbf{Frame Conditions (\texttt{assigns}):} The assigns clause solves the frame problem by listing exactly which memory locations may be modified. Every pointer‐ or array‐based region that the function writes to must appear here; anything not listed is guaranteed to remain unchanged. Scalar locals and by‐value parameters are implicitly excluded.

    \item \textbf{Loop Contracts \cite{Ernst2022}:}  When a function contains a \texttt{for}, \texttt{while}, or \texttt{do\ ...\ while} loop, ACSL provides \emph{loop contracts} to support both partial correctness and termination. Loop contracts include (1) \textbf{Loop invariants:} An assertion that must hold \emph{before and after} each iteration.  It captures the inductive property that lets the verifier prove the loop body preserves a key property. (2) \textbf{Loop variants:} A non-negative integer expression that strictly decreases on every iteration, ensuring the loop cannot run forever. (3) \textbf{Loop assigns:} A frame condition for the loop body, listing exactly which variables or memory regions may be modified in each iteration.

\end{itemize}

\subsection{Verification of ACSL Specifications Using Frama-C}

Frama-C \cite{frama-c} is an open-source platform for static analysis and formal verification of C programs. Its modular architecture enables various plug-ins to operate collaboratively. Among these, the \textbf{WP} (Weakest Precondition) plug-in \cite{frama-c-wp} serves as the core tool for deductive program verification. WP uses a Hoare logic weakest-precondition calculus to generate verification conditions (VCs) that capture the correctness of the code with respect to the specifications. In essence, WP computes logical conditions that must hold for the code to satisfy all its ACSL annotations. These conditions are then discharged by automated theorem provers (SMT solvers such as Alt-Ergo or Z3) or sent to interactive proof assistants (e.g. Coq) to attempt a proof. If all VCs are proven, the implementation is considered formally verified against its ACSL specification. This approach enables rigorous verification: any violation of a precondition, postcondition, or invariant will result in an unproven VC, indicating a potential bug or an incomplete specification.

\subsection{Code Property Graphs}

CPG is a unified program representation that combines multiple code analysis models—such as the abstract syntax tree (AST), control flow graph (CFG), data dependency graph (DDG), and Control Dependency Graph (CDG) —into a single, extensible graph structure. The CPG enables comprehensive reasoning about both the syntactic and semantic aspects of source code. Each node in the graph represents program elements (e.g., expressions, statements, variables), while edges capture relationships such as control flow, data flow, and syntax hierarchy. In the context of specification synthesis, CPGs provide valuable semantic features, such as variable dependencies, and loop structures, that can guide LLMs in generating verifiable specifications.

The CPG of a C program is a triple $(V, E, L)$, where $V$ is a set of nodes, $E \subseteq V \times V$ is a set of edges and $L: E \rightarrow 2^T$ is a labeling function that maps edges to a subset of labels $T$=\{\textit{AST}, \textit{CFG}, \textit{DDG}, \textit{CDG}\}. We use $L(e)$ or $L(v_i, v_j)$ to denote the set of labels for edge $e$ or $(v_i, v_j)$. For example, $L(v_1, v_2)$ = $\{AST, DDG\}$ means that edge $(v_1, v_2)$ appears in both the AST and DDG representations. We denote the set of all AST edges as $E_{AST}$ = 
$\{(v_i, v_j) | AST \in L(v_i, v_j) \}$ and the set of all AST nodes as $N_{AST}$ = ${v_i, v_j | (v_i, v_j) \in E_{AST}}$. This is similar for CFG, DDG, and CDG.

\section{The AutoACSL Framework}

\begin{figure}[t]
  \centering
  \includegraphics[
    width=\linewidth,
    height=0.55\textheight,
    keepaspectratio,
    trim=0mm 0mm 2mm 1mm,
    clip=true
  ]{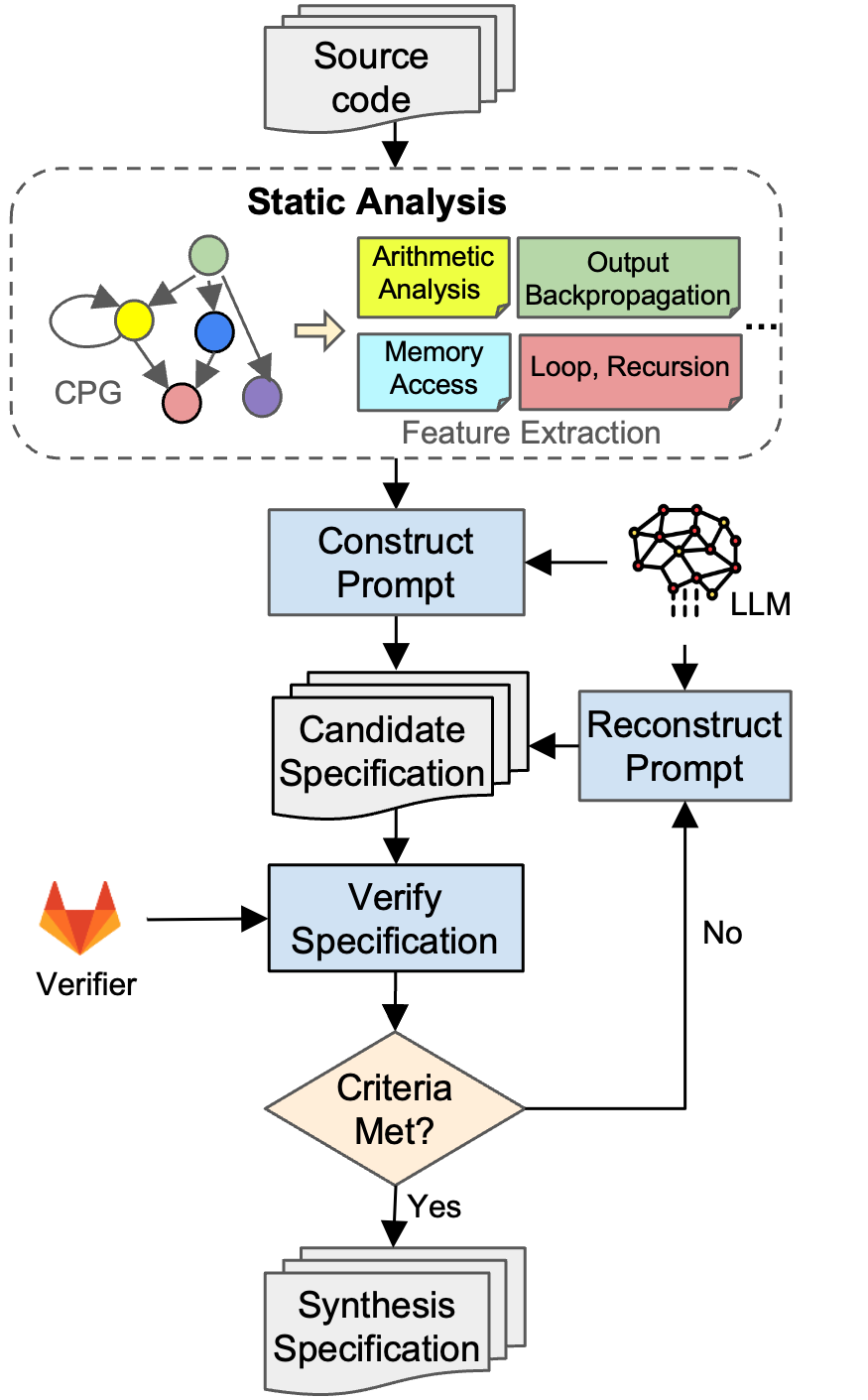}
  \caption{The AutoACSL Framework}
  \label{fig:Framework}
\end{figure}

As illustrated in Figure \ref{fig:Framework}, AutoACSL generates verifiable ACSL specifications from C code by integrating an LLM with insights from static analysis. Given a C program, AutoACSL first transforms the source code into a CPG and performs semantic analysis to extract key features—such as arithmetic operations, loop and recursion structures, and output backpropagation. These features provide essential context for guiding the synthesis of ACSL specification components, including preconditions, postconditions, frame conditions, loop contracts, and termination measures for recursion. The extracted features are then used to construct a structured prompt, which is provided to the LLM to generate a candidate specification alongside the original source code. This candidate specification is submitted to a verification tool—specifically, Frama-C/WP—to evaluate its correctness. The verification outcome falls into one of the following categories:
\begin{itemize}
\item \textbf{Syntax error}: The specification contains syntactic issues that prevent the verifier from generating verification conditions (VCs).
\item \textbf{Full proof}: All generated VCs are successfully discharged, confirming the specification's correctness.
\item \textbf{Partial proof}: Some VCs are verified, but one or more remain unproven. The proof ratio is defined as the number of discharged VCs divided by the total number of VCs.
\end{itemize}

If a candidate specification satisfies the predefined termination criteria, it is accepted as the final synthesized specification. Otherwise, the system reconstructs the prompt, incorporating feedback or refined inputs, and re-invokes the LLM to generate an improved candidate. This iterative feedback loop continues until one of the following termination conditions is met:

\begin{itemize}
\item The candidate specification achieves a full proof, meaning all VCs are successfully discharged.

\item There is no observable progress toward improving the specification or verification outcomes compared to previous iterations. We define the verification outcome of iteration $t$ by the counts of VC statuses returned by the verifier, including proved $ P_t $ , unknown $U_t$, failed $F_t$, and timeout $ T_t $. Because the total number of generated VCs may change across iterations when specifications are strengthened, we judge progress by changes in these outcomes rather than by VC count alone. We consider an iteration to make observable progress if it strictly improves the verification outcome, prioritizing reductions in $ F_t $, followed by an increase in $ P_t $  or a reduction in $ U_t $ + $ T_t $.

\item A timeout condition is reached. For example, the loop terminates if the candidate specification continues to exhibit syntax errors after a predefined number of iterations.
\end{itemize}

The progressive criterion described above does not guarantee that the current candidate is strictly better than previous ones. If the feedback loop terminates without producing a fully verified candidate specification, we select the one with the highest proof ratio as the synthesized specification. If no such candidate exists, preference is given to a syntax-error-free candidate over those containing syntax errors. If the termination criteria are not met, the candidate specification either contains syntax errors or includes one or more unproven VCs. To proceed with the next iteration of the loop, a new prompt is constructed by combining the current specification with the error messages or the unproved VCs returned by the verifier.

% \begin{figure*}[t]
%   \centering
%   \includegraphics[width=\textwidth]{framework.png}
%   \caption{The AutoACSL Framework}
%   \label{fig:Framework}
% \end{figure*}

\section{Feature Extraction from the CPG}

The feature extraction aims to provide the LLM with the critical contextual information needed to synthesize correct specifications.
The process begins with the identification of all arithmetic operations involving input variables, such as function parameters and global variables. This information is essential not only for generating precise preconditions, but also for understanding how inputs are transformed—knowledge that is later used to formulate postconditions. We then perform backward propagation analysis to reconstruct the expressions of output (e.g., return values and mutable variables) along all execution paths. These expressions directly inform the generation of postcondition clauses. In parallel, we conduct an in-depth analysis of loop structures, extracting their control variables, update rules, and termination conditions. This information is used to derive loop contracts as well as any additional preconditions required for loop correctness. Finally, we detect recursive call sites and infer the decreasing argument required to ensure termination, contributing to the completeness of the generated specifications. The feature extraction tasks described above involve a variety of algorithms. 

\begin{figure}[t]
  \centering
  \includegraphics[
    width=\linewidth,
    height=0.55\textheight,
    keepaspectratio,
    trim=0mm 0mm 2mm 1mm,
    clip=true
  ]{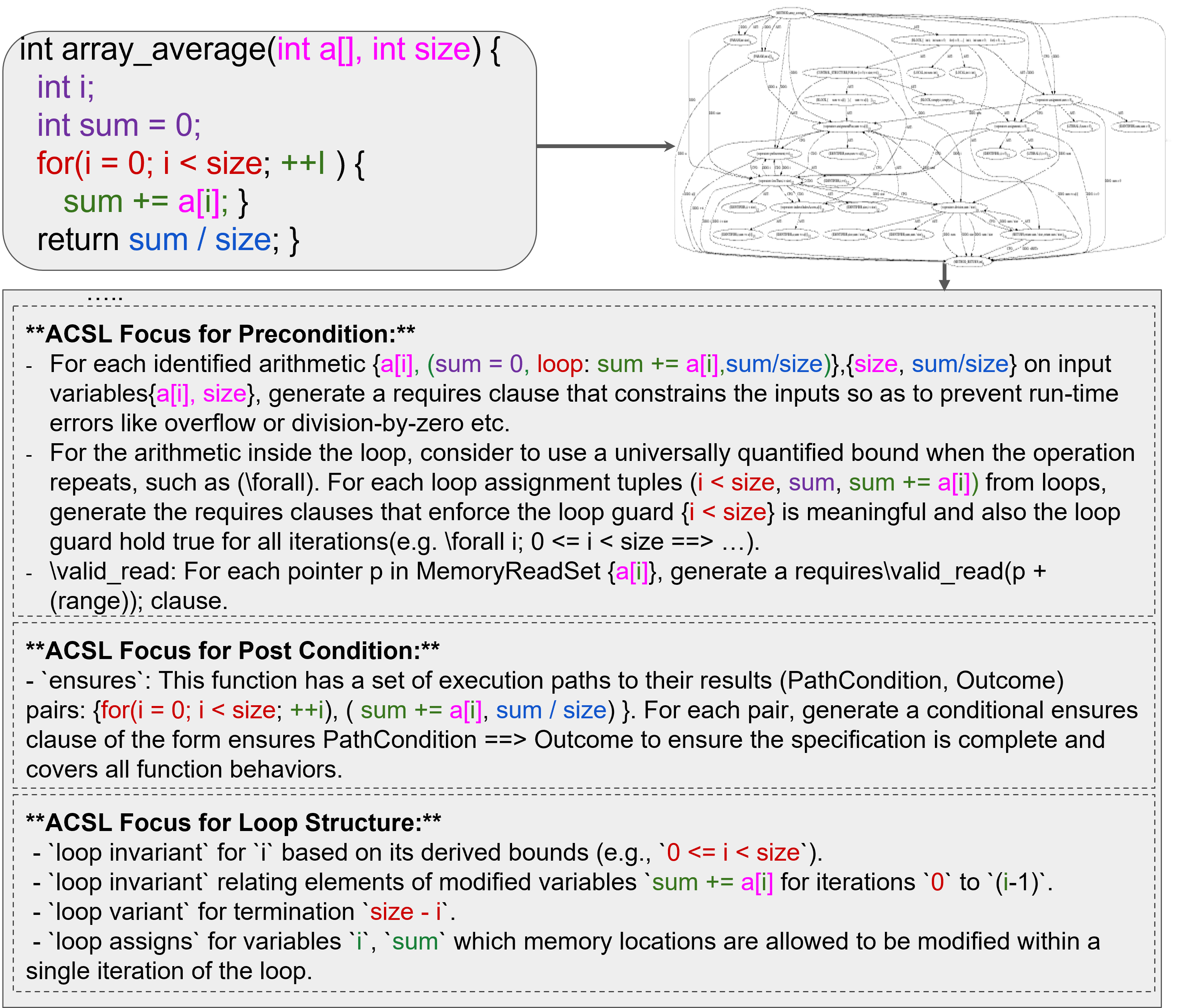}
  \caption{Sample code and prompt.}
  \label{fig:example}
\end{figure}

For illustration purposes, Figure \ref{fig:example} presents a running example that includes both the source code and a portion of the prompt constructed from the corresponding CPG features. Each highlighted segment in the prompt can be traced back to a matching code snippet in the source, using consistent color coding: pink for input variables, purple for local variables, red for loop conditions, green for assignments, blue for return expressions, and so on.

\subsection{Analysis of Input Variables and Arithmetic Operations}

This analysis aims to construct a list of arithmetic operation sequences involving input variables, which are essential for the accurate formulation of pre- and postconditions. A sound and robust ACSL specification must ensure the absence of all undefined or unsafe runtime behaviors.
Consider the \texttt{array\_average} function shown in Figure \ref{fig:example}. If the preconditions are not properly specified, two potential sources of unsafe behavior may arise: (a) Integer overflow in the expression \texttt{sum += a[i]}, and (b) Division by zero in the expression \texttt{sum/size}.

\begin{algorithm} [!t]
\caption{Extracting Arithmetic Operations Related to Input Variables}
\label{alg:input-ddg}
\IncMargin{1em}
\SetNlSkip{0.4em}
\SetInd{0.3em}{0.8em}
\DontPrintSemicolon
\SetAlgoLined

\KwIn{CPG $\langle V,E,L \rangle$, BackwardDDGPath()}
\KwOut{Input variables with their DDG operation sequences}

\KwSty{Parameter:}\\
\Indp
$L_p$, store input variables\\
$P$, store an input variable's operation sequences\\
$D$, store input variables with DDG operation sequences\\
\Indm

    Let $\varepsilon \in V$ where $type(\varepsilon) = \texttt{Method}$\;

    \tcp{Find input parameters and global variables}
    \ForEach{$(\varepsilon, \alpha) \in E_{AST}$}{
        \If{$type(\alpha) \in$ \{Param, Global\}}{
            Add $\alpha$ to $L_p$\;
        }
    }

    \ForEach{$p \in L_p$}{
        Initialize $P \gets \{[p]\}$\;
        \tcp{While the last node has DDG edges}
        \While{$\exists\, \pi \in P$ and $\exists (\pi[-1],u) \in E_{DDG}$}{
            $P' \gets \emptyset$\tcp*{store a temporary sequence}

            \ForEach{$\pi \in P$}{
                $v \gets \pi[-1]$\tcp*{$v$ is the last node in $\pi$}

                \ForEach{$(v,u) \in E_{DDG}$ and $u \notin \pi$}{
                    $E_u \gets \{ e ~|~ (u,e) \in E_{AST} \}$\;
                     \tcp*{If there are new variables exist}
                    \If{$\exists\, e \in E_u$ where $type(e) = \texttt{Identifier}$ and $e \neq p$}{

                        Add a new sequence (concatenating $\pi$ and $BackwardDDGPath(e)$) to $P'$

                    }\Else{
                        Add a new sequence (concatenating $\pi$ and $u$) to $P'$
                    }
                }
            }

            $P \gets P'$\;
        }

        $D[p] \gets P$\;
    }

    \KwRet $D$\;

\end{algorithm}

To address such risks, our approach constructs complete data dependency paths for each input variable to capture the arithmetic operations that define their computational influence throughout the code. We first identify the set of input variables $L_p$. These include both formal parameters and global variables accessible to the function. To collect them, we locate the method declaration node $\varepsilon$ and inspect all AST children connected via $(\varepsilon, \alpha) \in E_{AST}$. If a child node $\alpha$ has a type in \texttt{\{Param, Global\}}, it is added to the input list $L_p$. For each input variable $p \in L_p$, we initialize a path set $P$ with the singleton sequence $[p]$. It then performs a fixed-point iterative expansion of paths using DDG edges. For each current path $\pi \in P$, the last node $v$ in the path is checked for outgoing DDG edges to successors $u$. If $u$ introduces new identifiers in its AST subtree—i.e., it contains an \texttt{Identifier} node $e$ that is different from $p$—we compute the full backward dependency path from $e$ via the \texttt{BackwardDDGPath()} function and append it to the current path. Otherwise, $u$ is appended directly. This exploration continues until no further valid DDG edges remain for expansion. Once all paths for input $p$ have been explored, the complete set of arithmetic dependency sequences is stored in $D[p]$. The algorithm returns the mapping $D$, which records how each input variable contributes to downstream computations through DDG-based dependency chains. 

It is important to note that, without arithmetic operation analysis, an LLM tasked with generating specifications directly from raw code may omit critical safety checks. For instance, in the \texttt{array\_average} function shown in Figure \ref{fig:example}, prompting the LLM with a generic instruction such as “write a complete ACSL contract” resulted in a specification that included a division-by-zero guard (requires \texttt{size > 0;}) but entirely omitted any condition to prevent integer overflow. When this incomplete specification was verified using Frama-C, an unproven VC related to potential overflow was raised. In contrast, incorporating arithmetic operation analysis transforms the task of specification synthesis from an under-constrained generation problem into a semantics-guided process, ensuring that such safety conditions are explicitly captured. By enhancing the prompt with arithmetic features, the LLM is able to generate a specification that includes the necessary bound to eliminate the integer overflow risk, as shown below:
\begin{quote}
\scriptsize
\noindent\texttt{/*@}\\
\texttt{requires size > 0;}\\
\texttt{requires \textbackslash forall integer k; 0 <= k < size ==>}\\
\hspace*{1.5em}\texttt{SPEC\_INT\_MIN <= \textbackslash sum(integer j, 0, k, a[j])}\\
\hspace*{3.0em}\texttt{<= SPEC\_INT\_MAX;}\\
\texttt{...}\\
\texttt{@*/}
\end{quote}

Algorithm 1 describes the process of extracting arithmetic operations for input variables.

\subsection{Analysis of Output  Backpropagation}
\begin{algorithm} [!t]
\IncMargin{1em}
\SetNlSkip{0.4em}
\SetInd{0.3em}{0.8em}

\caption{Extracting Arithmetic Operations Related to Output Variables}
\label{alg:output-ddg1}
\DontPrintSemicolon
\SetAlgoLined
\KwIn{CPG $\langle V, E, L \rangle$, input varialbe list $L_p$, BackwardDDGPath()}
\KwOut{Local output variables with their DDG sequences}

\KwSty{Parameter:}\\
\Indp
$L_r$, store local variables in return expressions \\
$R$, store a local output variable's operation sequences\\
$D$, store input variables with DDG operation sequences\\
\Indm

\ForEach{$n \in V$ and $type(n) = \texttt{Return}$}{ 
  \If{$\exists (n, r) \in E_{AST}$ where $type(r) = \texttt{Identifier}$}{
    \ForEach{$(n,r) \in E_{AST}$ where $type(r) = \texttt{Identifier}$ and $r \notin L_p$}{
      Add $r$ to $L_r$
    }  
  }
  \Else{
    $D[n] \gets \{ [n] \}$\tcp*{Return literal}
  }
}

\ForEach{$r \in L_r$}{
 
    Initialize $R \gets \{ [r] \}$\;

    \While{$\exists\, \pi \in R$, $\exists (v,\pi[0]) \in E_{DDG}$ and $v \notin \pi$}{
      Initialize $R' \gets \emptyset$

      \ForEach{$\pi \in R$}{
        Let $u \gets \pi[0]$\tcp*{$u$ is the first node in $\pi$}

        \ForEach{$(v,u) \in E_{DDG}$ and $v \notin \pi$}{
          $E_u \gets \{ e ~|~ (v,e) \in E_{AST} \}$ \tcp*{If there are new variables exist}
                    \If{$\exists\, e \in E_u$ where $type(e) = \texttt{Identifier}$ and $e \neq r$}{

                        Add a new sequence (concatenating $BackwardDDGPath(e)$ and $\pi$) to $R'$

                    }\Else{
                        Add a new sequence (concatenating $v$ and $\pi$) to $P'$
                    }

          % Check CFG condition
          % \If{$\exists (w,v) \in E_{CFG}$ and $type(w) = \texttt{Control}$}{
          %   $R' \gets  R' \cup \{[(w)] + [v] + \pi\}$\; \tcp*{Insert condition before v}
          % }\Else{
          %   $R' \gets R' \cup \{[v] + \pi\}$\;
          % }
        }
      }
      $R \gets R'$\;
    }
    $D[r] \gets R$

}

\KwRet $D$\;
\end{algorithm}
Postconditions typically describe the behavior of a function’s return value and the final state of any mutable variables, such as the memory referenced by pointers or the values of global variables. Without a structured backpropagation analysis of return expressions and mutable variable updates, an LLM may struggle to correctly infer the mathematical structure of loop-based accumulations or to accurately handle multi-path return behaviors.

To support postcondition synthesis, we extract the data dependency sequences that compute output values, particularly focusing on return values derived from local variables. We begin by identifying local variables that contribute to return expressions. For each \texttt{Return} node $n$ in the CPG $G = \langle V, E, L \rangle$, if there exists an AST child $r$ of type \texttt{Identifier} and $r$ is not in the known input variable list $L_p$, we record $r$ as a return-related local output variable and add it to the list $L_r$. If the return statement directly returns a literal (i.e., no identifier is found), we store a singleton path $[n]$ in the result mapping $D[n]$ to capture this direct value flow. 

Next, for each local return variable $r \in L_r$, we initialize a set $R$ to store backward data dependency paths starting from $r$. We iteratively expand these paths using backward DDG traversal. For each current path $\pi \in R$, we examine all valid DDG edges $(v, u)$ where $u$ is the current head of $\pi$. For each such edge, we determine whether $v$ introduces new variables by checking its outgoing AST edges. If $v$ leads to an \texttt{Identifier} node $e$ not equal to $r$, we invoke the \texttt{BackwardDDGPath} function on $e$ to reconstruct its dependency history, and prepend it to the current path $\pi$. Otherwise, we directly prepend $v$ to $\pi$. The new paths are accumulated in $R'$ and replace $R$ for the next iteration. This process continues until no further expansions are possible.

Once the DDG expansion stabilizes for a return variable $r$, the set $R$ is stored in the mapping $D[r]$. After all return variables have been processed, the algorithm returns the complete mapping $D$, which captures the full backward arithmetic computation chain for each return-derived output value.

The result $D$ of this analysis is a set of (PathCondition, ReturnExpression) pairs, each representing a condition under which a specific return expression is produced. These pairs are directly translated into corresponding ACSL ensures clauses via prompt-based specification synthesis. Algorithm 2 describes the process.

\subsection{Analysis of Loop Structure}
Loop analysis aims to extract semantic features necessary for generating a complete loop contract, including the loop assigns, loop invariant, and loop variant clauses. A secondary objective is to provide loop-related information that supports the construction of the function’s overall precondition and postcondition.

We represent loop-related features with  conditional assignment tuples in the form (\texttt{C}, \texttt{T}, {U}), where 
\texttt{C} refers to the loop control condition, \texttt{T} is a target mutable variable in the loop, and \texttt{U} is a sequence of assignments that update the target variable in the loop.
For the \texttt{array\_average} example, we have two tuples:
(1) $(0 < i < size,\ i,\ i + 1)$ from \texttt{i++} operation, (2) $(0 < i < size,\ sum,\ sum + a[i])$ from \texttt{sum += a[i]} operation.

We obtain the $(C, T, U)$ tuples for each loop in three steps.  First, we locate the \texttt{CONTROL} node in the CPG and traverse the CDG from the loop entry point to collect all branching conditions. These conditions are composed into the loop control condition $C$. Second, within the AST subtree governed by each control branch, we identify all assignment statements and extract the left-hand side variables that are mutated in the loop body. Each variable is considered a target $T$. Third, for each target variable $T$, we perform a backward traversal on the DDG to collect all update operations that influence $T$ within the loop. The resulting sequence of dependent assignments constitutes the update component $U$ of the tuple.

To synthesize the \texttt{loop variant} clause, we distinguish between conditional and unconditional loops. For conditional loops, such as \texttt{for (i=0; i<size; ++i)}, the termination condition is derived directly from the loop control condition. The expression \texttt{size-i} is synthesized as a strictly decreasing variant that ensures loop termination. For unconditional loops (e.g., \texttt{while(1)}), where termination is not structurally guaranteed, we perform a CDG-based analysis to identify internal exit paths such as \texttt{break} or \texttt{return}. For each path, we locate its guarding condition and synthesize a variant expression based on the distance to this condition. If the analysis detects no reachable exits, the loop is correctly classified as intentionally infinite, and no \texttt{loop variant} is produced.

As mentioned before, the extracted loop features also support the generation of the function contract. For example, the combination of memory access \texttt{a[i]} and loop condition \texttt{i < size} reveals a dependency on the \texttt{size} parameter. To ensure that the access range \texttt{0..size-1} is valid, it must hold that \texttt{size > 0}, which leads to the LLM generating the corresponding function precondition: \texttt{requires size > 0}. Moreover, the loop features $(C, T, U)$  aid the output variable backpropagation analysis, enabling the generation of precise postconditions that reflect the loop’s cumulative effects.

\subsection{Analysis of Recursive Structures}

Proving termination is a critical component in the verification of recursive functions. In ACSL, this is expressed through the \texttt{decreases} clause, which defines a non-negative measure that must strictly decrease with each recursive call. Inferring a valid termination measure requires global reasoning about the function's call structure and parameter transformations—something that is often difficult for an LLM to deduce accurately from code patterns alone.

To address this issue, AutoACSL analyzes the AST to identify all recursive call sites and examine the arguments passed at each invocation. Additionally, it leverages the DDG to determine which parameters consistently decrease in value across recursive calls. These parameters are then designated as the formal termination measure. By extracting and explicitly supplying this measure—and recognizing that it must be non-negative at the entry point — our prompt construction provides the LLM with the precise context needed to generate both the \texttt{decreases} clause and a corresponding \texttt{requires} clause that enforces the non-negativity constraint.

\begin{table*}[t]
\centering
\small
\setlength{\tabcolsep}{4pt}
\renewcommand{\arraystretch}{1.15}
\caption{Experimental Results }
\begin{tabular}{lccccccccccccccc}
\toprule
\textbf{Model} &
\multicolumn{3}{c}{\textbf{Overall}} &
\multicolumn{3}{c}{\textbf{DS-I}} &
\multicolumn{3}{c}{\textbf{DS-II}} &
\multicolumn{3}{c}{\textbf{DS-III}} &
\multicolumn{3}{c}{\textbf{DS-IV}} \\
\cmidrule(lr){2-4}\cmidrule(lr){5-7}\cmidrule(lr){8-10}\cmidrule(lr){11-13}\cmidrule(lr){14-16}
&
GSR & FPR & AVC &
GSR & FPR & AVC &
GSR & FPR & AVC &
GSR & FPR & AVC &
GSR & FPR & AVC   \\
\midrule

GPT-o4 Mini  &92\%  &88\% &14 &90\%  &84\%  &19  &96\%  &84\%  &\underline{15}  &90\%  &90\%  &17  &92\%  &89\% &12  \\
GPT-5.2      &92\%  &88\% &\underline{17} &85\%  &86\%  &\underline{21}  &93\%  &\underline{96\%}  &14  &\underline{100\%}  &89\%  &\underline{22}  &91\%  &87\% &\underline{16}  \\
Grok-4.1     &74\%  &91\% &12 &56\%  &78\%  &15  &88\%  &92\%  &14  &87\%  &89\%  &18  &83\%  &92\% &11  \\
Gemini-3   &\underline{98\%}  &\underline{96\%} &16
&\underline{100\%}  &\underline{88\%}  &20  &\underline{100\%}  &95\% &\underline{15}  &99\%  &\underline{95\%}  &21  &\underline{97\%}  &\underline{98\%} &15  \\
\bottomrule
\end{tabular}
\label{tab:model_by_dataset_metrics}
\end{table*}
\section{Prompt Construction}
The goal of prompt construction in AutoACSL is to incorporate semantic features extracted from the CPG into structured prompts that guide specification synthesis. These prompts act as a semantic blueprint, steering the model toward generating precise and context-aware ACSL clauses rather than relying solely on surface-level patterns in the source code. Prompt construction is organized into five dedicated modules, each corresponding to a core component of an ACSL specification: preconditions, frame conditions, postconditions, loop contracts, and termination conditions for recursive functions.

Each module contributes guidance only when the analyzed function exhibits the corresponding program feature. If a feature is absent, the associated prompt module is omitted to avoid introducing irrelevant or misleading constraints. For example, loop-related or recursion-related guidance is included only when loops or recursive calls are present in the code.

\emph{Precondition Prompt Construction.} To generate ACSL requires clauses, AutoACSL incorporates four feature categories, including
arithmetic operation sequences discussed in Section 5.1, loop-related features for quantified constraints discussed in Section 5.3, and recursive termination measures. The following are sample prompt patterns. 

\begin{itemize}

\item Arithmetic operation sequences:   ``For each recorded arithmetic expression {arithmetic sequence set} on variables {input\_variable\_1, input\_variable\_2, ...}, generate an ACSL requires clause that constrains the inputs to prevent runtime errors like overflow or division-by-zero. For arithmetic inside loops, use a universally quantified bound, such as $ \forall k;\ \ldots $."

 \item Loop-related features for quantified constraints:  
``For each loop assignment tuple {(Condition, Target, Update)}, generate requires clauses that ensure the loop guard Condition is meaningful and holds for all iterations, such as $ \forall i;\ 0 \leq i < size\ \Rightarrow\ \ldots $."

\item Recursive termination measures:   
``Since the function recurses on {m}, add requires $ m \geq 0 $ to ensure a valid non-negative termination measure."

\end{itemize}

\emph{Frame Condition Prompt Construction.} To generate the \texttt{assigns} clause, AutoACSL uses the \texttt{MemoryWriteSet}. The prompt is:  
``Each memory region in {MemoryWriteSet} may be modified by the function. Generate an ACSL assigns clause listing each written memory location."

\emph{Postcondition Prompt Construction.} AutoACSL collects (PathCondition, Outcome) pairs for return expressions and mutable variables (Section~5.2,). These are translated into the following prompt:  
``For each (PathCondition, Outcome) pair, generate a conditional postcondition of the form: ensures PathCondition $ \Rightarrow $ Outcome."

\emph{Loop Contract Prompt Construction.} Loop contract prompts are constructed using the (\texttt{C}, \texttt{T}, \texttt{U}) tuples discussed in Section~5.3:  
``Generate a loop invariant that describes how {Target} evolves based on the update rule {Update}, under the condition {Condition}. The invariant must also ensure the condition is meaningful." In addition, the prompt includes loop-specific frame and termination clauses:  
``Loop writes: {LoopAssignsSet}. Generate a loop assigns clause."  
``If a LoopVariant is available, generate a loop variant clause for termination using: {LoopVariant}." If the loop is classified as intentionally infinite, the prompt explicitly omits the variant clause.

\emph{Recursive Termination Prompt Construction.} For recursive functions, AutoACSL uses decreasing argument analysis introduced in Section~5.4. The prompt includes:  ``Add a decreases clause using the identified measure."  ``Also add a requires clause ensuring the measure is non-negative."

\section{Experiments}

Our experiments aim to address the following questions: \textbf{(RQ1)}: How effective is AutoACSL in generating verifiable ACSL specifications from C programs? \textbf{(RQ2)}: How does the CPG-based analysis and prompting contribute to the effectiveness of AutoACSL?
\textbf{(RQ3)}: How does AutoACSL’s performance compare to existing state-of-the-art research for specification synthesis from C code?

\subsection{Experiment Setup}
\paragraph{\textbf{Environment.}}
AutoACSL is implemented on top of Frama-C version 30.0 (Zinc), using the WP to verify generated ACSL specifications against the corresponding
C programs. While AutoACSL is not tied to a specific LLM, our experiments focus on four representative LLMs with different sizes and capabilities:
GPT-5.2, GPT-o4 mini, Gemini-3, and Grok-4.1-fast-reasoning.

\paragraph{\textbf{Subject Programs.}}
We use a total of 604 C programs drawn from four datasets. The first dataset (DS-I) consists of 41  programs collected from GitHub, covering basic arithmetic, array manipulation, and control-flow patterns. The second dataset (DS-II) includes 57 programs from a widely used benchmark for specification synthesis \cite{Cheng2024}. The availability of detailed verification results for these programs enables direct comparison of AutoACSL with prior work.

The third dataset (DS-III) consists of 93 programs from another benchmark \cite{Cheng2024}, excluding those with undefined function calls, as such programs cannot be fully verified. The fourth dataset (DS-IV) contains 413 programs from the CASP repository \cite{Hertzberg2025}, which is specifically designed for formal verification tasks and includes a wide range of memory- and loop-intensive code. Collectively, these programs span a broad spectrum of characteristics, including varied use of arrays, pointers, loops, and nested control structures. This diversity enables a comprehensive assessment of both the robustness and scalability of AutoACSL.

\paragraph{\textbf{Performance Metrics.}}
We measure the effectiveness of specification synthesis using the following primary metrics: Generation Success Ratio (GSR), Full Proof Ratio (FPR), and Average Number of Verification Conditions (AVC). GSR is defined as the ratio of subject programs for which the generated specifications contain no syntax errors to the total number of subject programs. FPR (or PPR) denotes the ratio of subject programs for which the generated specifications are fully (or partially) proved. AVC represents the average number of verification conditions associated with all generated specifications that are fully or partially proved. A higher AVC may indicate more complex programs or more comprehensive specifications.

\subsection{RQ1: Effectiveness of AutoACSL}

Table \ref{tab:model_by_dataset_metrics} summarizes the experimental results of AutoACSL across four datasets and four LLMs. Overall, AutoACSL achieves consistently strong performance across models, with Gemini-3 delivering the best results, reaching 98\% GSR, 96\% FPR, and an average of 16 verification conditions, indicating that it generates both highly verifiable and sufficiently expressive specifications. GPT-o4-mini and GPT-5.2 also perform robustly, both exceeding 92\% GSR and 88\% FPR, while producing 14–17 AVC, demonstrating that AutoACSL remains effective even with smaller or less capable models. In contrast, Grok-4.1 exhibits lower generation robustness (74\% GSR overall) and fewer verification conditions (12 AVC); however, its FPR remains competitive once specifications are successfully generated, suggesting that its primary limitation lies in specification generation robustness rather than verification strength.

Across datasets, AutoACSL maintains stable verification effectiveness while consistently generating non-trivial specifications over diverse program collections with varying code styles, control structures, and memory usage patterns. On DS-I and DS-II, which primarily consist of smaller programs with simpler arithmetic and control structures, AutoACSL achieves FPR of up to 88\% and 96\%, respectively, with AVC values reaching 20 and 15, indicating expressive yet verifiable contracts. Performance remains robust on DS-III and DS-IV, which include programs with heavier use of arrays, pointers, loops, and nested control flow, where Gemini-3 attains near-perfect GSR (99\%) and high FPR (95–98\%) while maintaining 15–21 AVC. Higher AVC values consistently align with strong FPR across all datasets, suggesting that AutoACSL improves specification completeness without sacrificing verifiability as program complexity increases.

\subsection{RQ2: Contribution of CPG-Guided Prompting}

\begin{figure}[t]
\centering
\includegraphics[width=0.72\columnwidth]{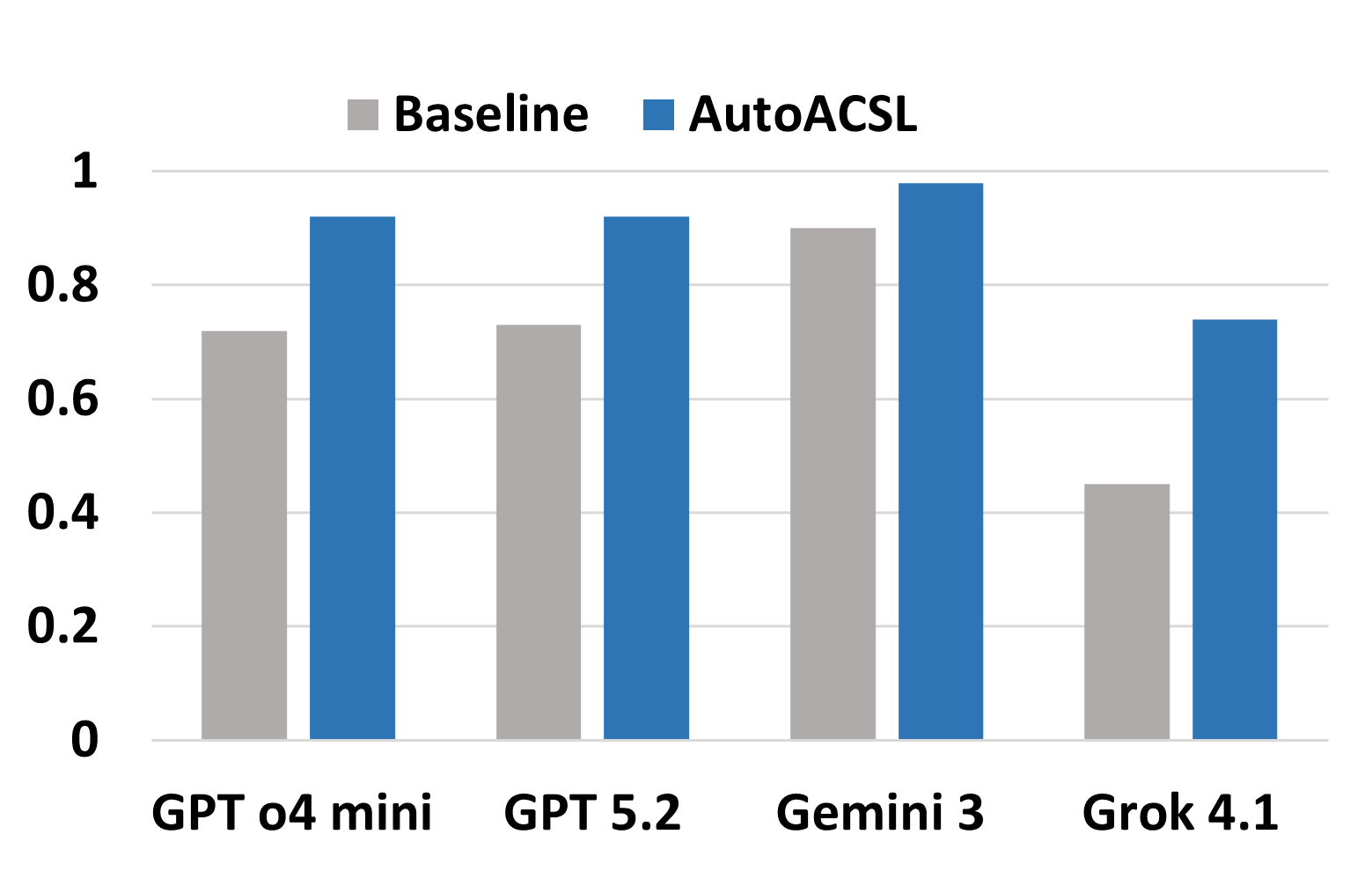}

{\small (a) Generation Success Ratios}

\vspace{0.4em}
\includegraphics[width=0.72\columnwidth]{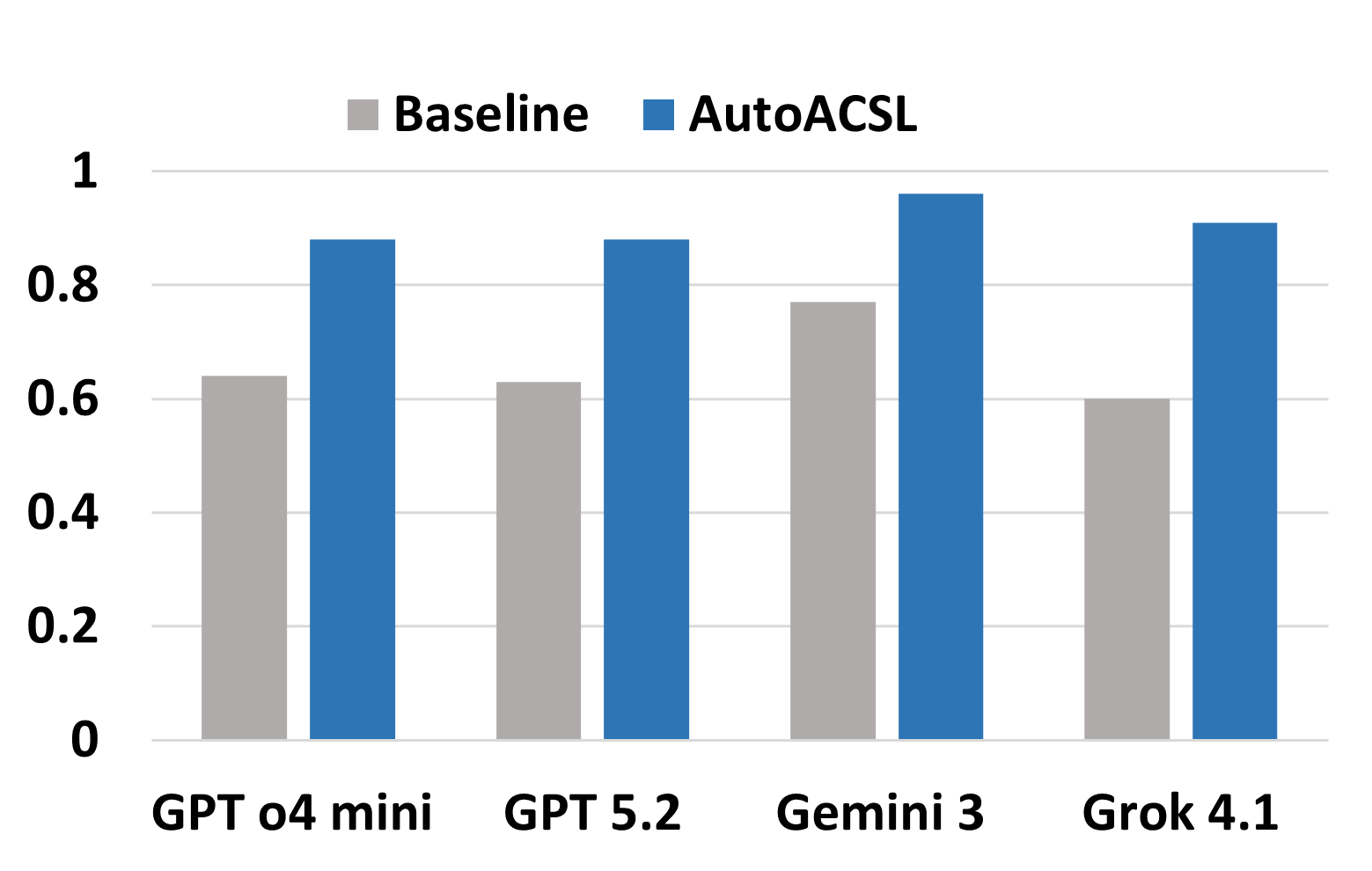}

{\small (b) Full Proof Ratios}

\caption{Results of Ablation Study}
\label{fig:Ablation}
\end{figure}

The performance of AutoACSL depends on both the underlying LLM and the CPG-guided prompting. To isolate the contribution of CPG-based analysis, we conduct an ablation study comparing AutoACSL with a baseline that relies solely on code-based prompting. Fig.\ref{fig:Ablation} shows the ablation results in terms of GSR and FPR. Overall, AutoACSL consistently outperforms the baseline on both metrics. Across all models, AutoACSL improves GSR by 8.9\%–64.4\% and increases FPR by 24.7\%–51.7\%, demonstrating that CPG-guided prompting significantly enhances both generation robustness and verification effectiveness.

Looking at individual models, the improvements in GSR are more pronounced for GPT-o4-mini, GPT-5.2, and Grok-4.1, where code-only prompting is more susceptible to syntax errors and incomplete specifications. In particular, Grok-4.1 exhibits the largest gain, with a 64.4\% relative improvement in GSR, indicating that semantic guidance plays a crucial role when the underlying model’s reasoning capability is limited. These gains in generation robustness are accompanied by substantial improvements in FPR, with increases of up to 51.7\%, showing that better-formed specifications are also more likely to be fully verified.

In contrast, Gemini-3 already achieves strong performance under the baseline setting. Nevertheless, AutoACSL still improves its GSR from 90\% to a near-perfect 98\%, along with a 24.7\% increase in FPR, indicating that CPG-based semantic features provide complementary benefits even when the base prompt is highly effective. Together, these results show that AutoACSL improves both metrics consistently, while the magnitude of improvement depends on the baseline capability of the underlying LLM.

\subsection{RQ3: Comparison with AutoSpec}

AutoSpec~\cite{Cheng2024} is a tool that leverages LLMs to generate specifications for C programs and, to the best of our knowledge, is the only tool directly comparable to AutoACSL. AutoSpec builds a call graph in which functions and loops are represented as nodes and employs an iterative process that refines LLM-generated specification candidates based on verification feedback.
For a fair comparison, we evaluate both tools on DS-II (the only dataset for which AutoSpec reports detailed per-program verification outcomes) and use GPT-3.5-turbo, the best-performing LLM reported in AutoSpec. On this dataset, AutoACSL fully verifies 45 out of 57 programs, whereas AutoSpec fully verifies 37 out of 57. Consequently, AutoACSL with GPT-3.5-turbo outperforms AutoSpec with GPT-3.5-turbo by 14\%.

\subsection{Threats to Validity}

We discuss the main threats to validity that might affect the experimental findings. 

\textbf{Internal Validity.}
Data memorization is a potential threat because the LLMs are trained on large-scale public datasets that may contain some benchmark programs. While the model could recall memorized content. AutoACSL does not rely solely on the source code. It combines code with semantic features extracted from static analysis. To evaluate whether memorization affects performance, we conducted ablation experiments comparing code-only prompts with our full pipeline. The results showed that prompts containing only code performed significantly worse, indicating that the model's success is not attributable to memorization but rather to structured guidance. Another internal threat is the LLM’s ability to interpret the generated prompts accurately. Despite systematic construction of prompts using semantic information from CPGs, edge cases may still be misinterpreted. We address this with iterative verification and manual inspection of sampled outputs to ensure correctness and consistency. Additionally, although our static analysis handles various features, it may miss critical patterns in particularly complex C constructs, such as function pointers or inline assembly. Nonetheless, empirical results across various programs—including cases where baseline tools fail—demonstrate the robustness of our feature extraction and generation approach.

\textbf{External Validity.}
AutoACSL is currently designed for C programs and ACSL annotations, which limits its applicability to other languages or formal specification systems. While the framework could potentially be adapted to languages like C++ or Rust, doing so would require significant changes in both feature extraction and prompt formulation. Another limitation is the dependency on a specific LLM version; future versions of LLMs may behave differently, which could affect reproducibility and consistency.

\section{Conclusions}

We have introduced AutoACSL, a novel framework that combines LLMs with CPG-based static analysis to automatically generate ACSL specifications for C programs. AutoACSL augments LLM inputs with structured semantic features, such as memory access patterns, arithmetic operations, control flow structures, and return value dependencies, which collectively form a semantic blueprint to guide the generation of precise and verifiable specifications, including constraints that mitigate potential runtime errors. Our empirical evaluation demonstrates that AutoACSL has achieved state-of-the-art performance.

Future work includes extending AutoACSL to support additional specification languages (e.g., JML or Rust’s Verus), incorporating fine-tuned LLMs trained on specification tasks, and applying the framework to industrial-scale software systems. AutoACSL provides a promising foundation for making formal specification authoring more accessible, accurate, and automated.

%\clearpage
\bibliographystyle{IEEEtran}
\bibliography{reference}

\clearpage
\appendix 

%\textbf{Appendices}

\subsection{Analysis of Memory Access}

The memory access analysis serves two purposes: (1) guiding the generation of memory safety preconditions such as \texttt{requires \textbackslash valid}, \texttt{\textbackslash valid\_read}, and \texttt{\textbackslash separated}; and (2) supporting the generation of frame conditions via the \texttt{assigns} clause. The analysis covers memory accesses through parameter-passed pointers and arrays, as well as reads and writes to global variables. Each access is classified as a read or write, and alias analysis is performed to detect overlapping memory regions. For example, accessing \texttt{a[i]} may cause a buffer overflow if \texttt{a} does not point to a valid memory region of sufficient size, requiring a precondition like \texttt{\textbackslash valid\_read}. Any memory location that is modified must also appear in the function’s \texttt{assigns} clause to ensure soundness.

To support the analysis, Algorithm~\ref{alg:mem-access} extracts three feature sets from the CPG: (1) \texttt{MemoryReadSet} includes all memory locations that are read by the function through pointer, array, or global variable accesses. (2) \texttt{MemoryWriteSet} consists of all memory locations that the function may write to during execution. (3) \texttt{AliasingSet} includes all pairs of input pointers or arrays that may access overlapping memory regions, where at least one is written to. In addition, the algorithm maintains a map $M$ that records the abstract memory locations accessed by each pointer or array variable.

\begin{algorithm}[!t]
\caption{Extracting Memory Access Features}
\label{alg:mem-access}
\DontPrintSemicolon
\SetAlgoLined
\KwIn{CPG $\langle V, E, L \rangle$}
\KwOut{\texttt{MemoryReadSet}($R$), \texttt{MemoryWriteSet} ($W$), \texttt{AliasingSet} ($A$), map from pointer to accessed abstract locations ($M$); \\
\KwSty{Parameters:} $L_m$: Memory-relevant variables; }

\ForEach{$v \in V$}{
  \If{$type(v) \in \{ \texttt{PARAM}, \texttt{GLOBAL} \}$ and $v \in$ \{Pointer, Array, DoublePointer\}}{
    Add $v$ to $L_m$; $M[v] \gets \emptyset$\;
  }
}

\ForEach{$m \in L_m$}{
  \ForEach{$(m, u) \in E_{AST}$}{
    \If{$type(u) = \texttt{CALL}$ and $u \in$ \{Dereference, Index, FieldAccess\}}{
      $\texttt{Access} \gets \texttt{READ}$\;
      \ForEach{$(u, a) \in E_{AST}$}{
        \If{$type(a) = \texttt{assignment}$ and $num(u) < num(a)$}{
            $\texttt{Access} \gets \texttt{WRITE}$\;
        }
      }

      \If{$type(m) = \texttt{DoublePointer}$}{
        Add $m$ to $R$, add $*m$ to $W$\;
        $M[m] \gets M[m] \cup \texttt{BackwardDDGPath}(m)$\;
        $M[*m] \gets M[*m] \cup \texttt{BackwardDDGPath}(*m)$\;
      }\Else{
        \If{$\texttt{Access} = \texttt{READ}$}{
          Add $m$ to $R$\;
        }\ElseIf{$\texttt{Access} = \texttt{WRITE}$}{
          Add $m$ to $W$\;
        }
        $M[m] \gets M[m] \cup \texttt{BackwardDDGPath}(m)$\;
      }
    }
  }
}

\ForEach{$(p, q) \in L_m \times L_m$ and $p \neq q$}{
  \If{$M[p] \cap M[q] \neq \emptyset$ and $(p \in W$ or $q \in W)$}{
    $A \gets A \cup \{(p, q)\}$
  }
}

\KwRet $R, W, A, M$\;

\end{algorithm}

Algorithm~\ref{alg:mem-access} begins by identifying all parameters of input pointer, array, or double-pointer types and globals (lines 1-5), which are collected into the memory-relevant variables set $L_m$. For each such variable, an empty abstract location set $M[v]$ is initialized. We then traverse the AST to identify dereference-style memory accesses such as pointer dereferencing, array indexing, and field access. Each access is classified as either a read or a write by analyzing its surrounding assignment structure. If the access appears on the left-hand side of an assignment (lines 11-12), it is labeled as a write; otherwise, it is a read (lines 8-9). For each access, the corresponding abstract memory location is computed by calling function  \texttt{BackwardDDGPath} (line 27). \texttt{BackwardDDGPath} performs a backward traversal along DDG edges to locate the nearest definition or input variable, thereby identifying the abstract memory location associated with the access. In the case of double pointers, both the pointer and its dereferenced value are handled separately (lines 17-18).

Each variable is added to the appropriate access set—$R$ for reads (line 22) and $W$ for writes (line 25)—and its access locations are stored in the map $M$. Finally, aliasing is detected by checking for overlapping access regions between distinct variable pairs (lines 32-36). If two variables share at least one accessed location and one of them is a write, the pair is added to the aliasing set $A$. 

Once we have obtained the three sets for the input pointers, input arrays, and global pointers, we guide the LLM to generate \texttt{requires \textbackslash valid} for each parameter in \texttt{MemoryWriteSet} to guard the memory write access, generate \texttt{\textbackslash valid\_read} for each parameter in the \texttt{MemoryReadset} to guard the read access, generate \texttt{requires \textbackslash separated} for aliasing pairs in \texttt{AliasingSet}. Memory safety preconditions for global array and scalar variable is unnecessary. A global array's base address is always valid because it's statically allocated and for global scalars like \texttt{int}, it can not be invalidly dereferenced. For all input pointers, arrays, global variables in \texttt{MemoryWriteSet}, we guide the LLM to generate correct and complete \texttt{assigns} clause. 

\section{Category-Wise Analysis with Respect to the Research Questions}

\subsection{Program Categories}

To better understand how structural complexity of programs affects the synthesis of verifiable specifications, we categorize subject programs into four groups based on their use of arrays, pointers, and loops. The categorization is derived from static analysis of each program and reflects increasing verification difficulty.

\begin{itemize}
  \item \textbf{C1: Scalar and Single-Loop Programs.} Programs that do not involve arrays or pointers and may contain at most a single loop. These programs typically involve simple arithmetic or conditional logic and pose minimal challenges for specification generation and verification.
  \item \textbf{C2: Multi-Loop and Single-Array Programs.} Programs that manipulate arrays and contain multiple loops, but do not involve pointer aliasing. This category introduces basic memory access patterns and loop reasoning but remains relatively straightforward due to the absence of aliasing. Verification in this category requires synthesizing quantified loop invariants and reasoning about array bounds and functional correctness across iterations. 
  \item \textbf{C3: Pointer-Centric Programs without Loops.}
  Programs that involve pointers or multiple arrays but contain no loops. The dominant verification difficulty arises from pointer aliasing and memory validity. Loop invariants are not required, but precise memory specifications are essential.
   \item \textbf{C4: Pointer- and Loop- Programs.}
  Programs that combine multiple loops with pointer-based or multi-array memory accesses. This category represents the most challenging verification setting, as it requires reasoning about quantified loop invariants together with pointer aliasing and memory framing. 
\end{itemize}
For programs containing multiple functions, we assign the category of the most complex function to represent the entire program.

\subsection{Category-Wise Analysis of Effectiveness (RQ1)}

Figure~\ref{fig:AblationCat} reports the GSR and FPR of AutoACSL across the four categories, evaluated using the best-performing model, Gemini-3.

AutoACSL achieves consistently high generation success ratios across all categories, reaching 99\% in C1, 96\% in C2, 98\% in C3, and 97\% in C4, which indicates that it reliably constructs syntactically valid ACSL specifications even as program structure becomes more complex. More importantly, AutoACSL attains strong full-proof performance on C2 and C3, with full proof ratios of 87\% and 95\%, respectively, covering two practically challenging verification settings. The high success in C2 suggests that AutoACSL can synthesize the quantified loop invariants needed to reason about iterative array computations, while the similarly high success in C3 shows that it can generate effective memory-safety and framing constraints. In contrast, C4 remains the hardest category: although generation remains robust at 97\%, the full proof ratio drops to 77\%, reflecting the difficulty of simultaneously discharging quantified loop obligations and precise aliasing and frame conditions in programs where loops and pointer-based updates interact. Overall, these results highlight AutoACSL’s effectiveness on non-trivial loop-centric and memory-centric verification tasks while identifying combined loop+pointer reasoning as an important direction for future improvement.

\subsection{Category-Wise Ablation Study of the CPG Analysis (RQ2)}

\begin{figure*}[h]
\centering
\begin{tabular}{cc}
\includegraphics[scale=0.65]{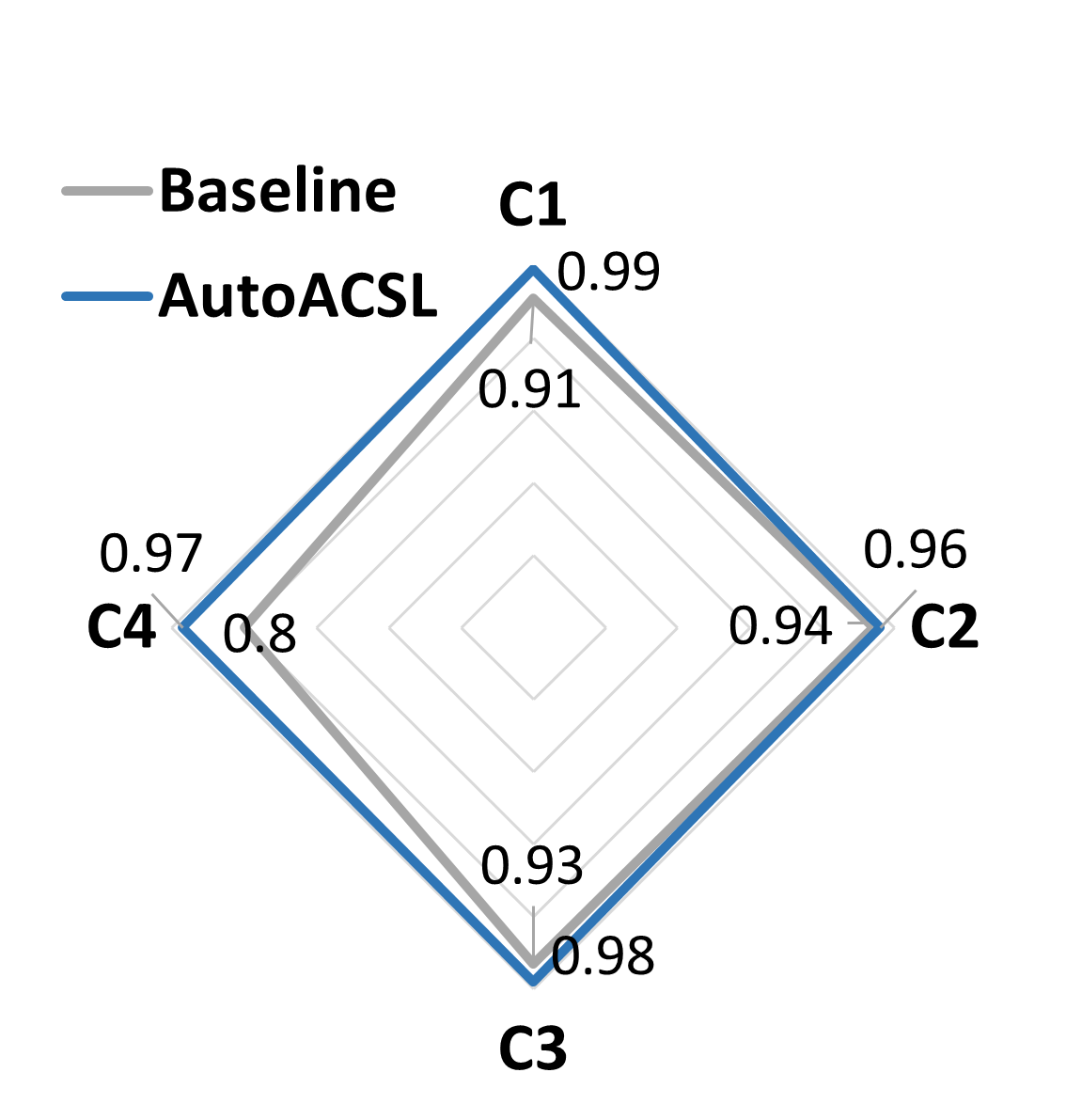}
&
\includegraphics[scale=0.65]{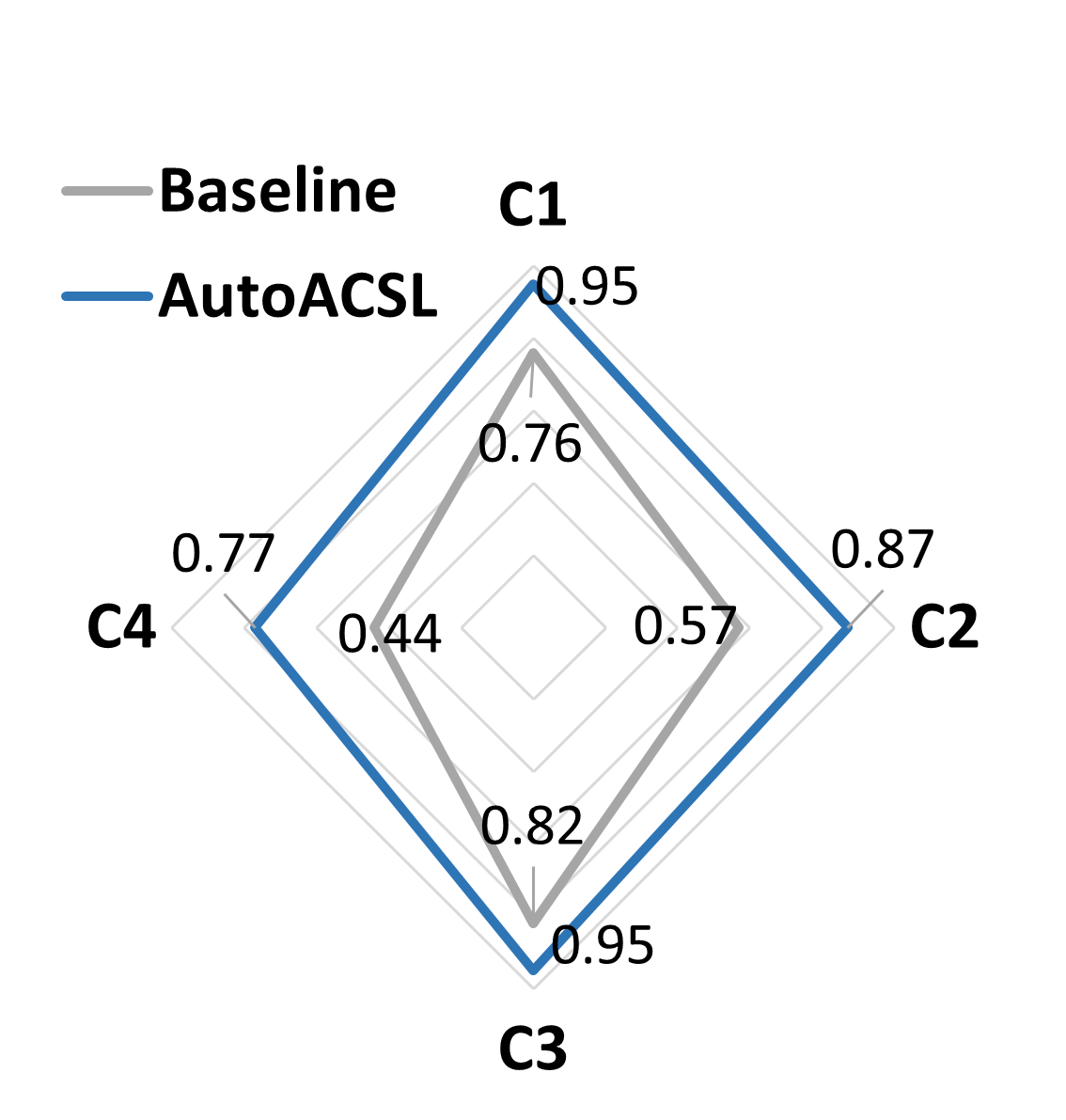}\\
(a) Generation Success Ratios  & (b) Full Proof Ratios \\

\end{tabular}
\caption{The Ablation Study on the Four Program Categories}
\label{fig:AblationCat}
\end{figure*}

Figure~\ref{fig:AblationCat} shows that AutoACSL consistently outperforms the baseline across all four program categories, with the performance gap widening as program structure becomes more complex. For simple programs without arrays or pointers (C1), both approaches achieve high verification success; nevertheless, AutoACSL improves the full passed ratio from 91\% to 99\%. In C2 programs containing a single array and loop, AutoACSL raises the full passed ratio from 76\% to 0.95\%, indicating more robust handling of loop and array semantics. The advantage becomes more pronounced for pointer- and array-intensive programs without loops (C3), where verification success improves from 57\% to 87\%. For the most complex category (C4), which combines arrays, pointers, and loops, AutoACSL achieves a full passed ratio of 77\%, substantially higher than the baseline (44\%) but still lower than in simpler categories. While semantic guidance significantly mitigates verification difficulty, jointly reasoning about complex memory interactions and loop behaviors remains challenging and leaves room for further improvement.

\subsection{Category-Wise Comparison with Existing Work (RQ3)}

We provide a category-wise comparison between AutoACSL with GPT-3.5-turbo and AutoSpec with GPT-3.5-turbo on the dataset DS-II. AutoACSL shows consistent progress across four categories. For C1, AutoACSL increases the full verification ratio by approximately 15\%, indicating improved robustness even for programs without complex memory usage or control-flow structure. For C2, the improvement is more pronounced, with the full verification ratio increasing by 19\%, highlighting AutoACSL’s stronger handling of loop invariants and array access constraints. For programs involving more complex memory usage (C3 and C4), AutoACSL achieves full verification on all programs in these categories, exceeding AutoSpec’s performance by approximately 13\%.

\balance

\end{document}